\documentclass[10pt,twocolumn,letterpaper]{article}

\usepackage[pagenumbers]{cvpr} % To force page numbers, e.g. for an arXiv version

\usepackage{mathtools}
\usepackage{graphicx}
\usepackage{multirow}
\definecolor{cvprblue}{rgb}{0.21,0.49,0.74}
\usepackage[pagebackref,breaklinks,colorlinks,allcolors=cvprblue]{hyperref}

\def\paperID{4757} % *** Enter the Paper ID here
\def\confName{CVPR}
\def\confYear{2025}

\title{SVFR: A Unified Framework for Generalized Video Face Restoration}

\author{Zhiyao Wang\textsuperscript{1,2}, Xu Chen\textsuperscript{2,*}, Chengming Xu\textsuperscript{2}, Junwei Zhu\textsuperscript{2†}, Xiaobin Hu\textsuperscript{2}, \\Jiangning Zhang\textsuperscript{2}, Chengjie Wang\textsuperscript{2}, Yuqi Liu\textsuperscript{1}, Yiyi Zhou$\textsuperscript{1†}$, Rongrong Ji$\textsuperscript{1}$ \\
\textsuperscript{1}Key Laboratory of Multimedia Trusted Perception and Efficient Computing,\\ Ministry of Education of China, Xiamen University, China \quad
\textsuperscript{2}Tencent Youtu Lab \\
}
\begin{document}
\maketitle

\begin{abstract}
    Face Restoration (FR) is a crucial area within image and video processing, focusing on reconstructing high-quality portraits from degraded inputs. Despite advancements in image FR, video FR remains relatively under-explored, primarily due to challenges related to temporal consistency, motion artifacts, and the limited availability of high-quality video data. Moreover, traditional face restoration typically prioritizes enhancing resolution and may not give as much consideration to related tasks such as facial colorization and inpainting. In this paper, we propose a novel approach for the Generalized Video Face Restoration (GVFR) task, which integrates video BFR, inpainting, and colorization tasks that we empirically show to benefit each other. We present a unified framework, termed as stable video face restoration (SVFR), which leverages the generative and motion priors of Stable Video Diffusion (SVD) and incorporates task-specific information through a unified face restoration framework. A learnable task embedding is introduced to enhance task identification. Meanwhile, a novel Unified Latent Regularization (ULR) is employed to encourage the shared feature representation learning among different subtasks. To further enhance the restoration quality and temporal stability, we introduce the facial prior learning and the self-referred refinement as auxiliary strategies used for both training and inference. The proposed framework effectively combines the complementary strengths of these tasks, enhancing temporal coherence and achieving superior restoration quality. This work advances the state-of-the-art in video FR and establishes a new paradigm for generalized video face restoration. Code and video demo are available at \url{https://github.com/wangzhiyaoo/SVFR.git}.
\end{abstract}

% \iffalse
\label{sec:methods}
\begin{figure}[t]
  \centering
   \includegraphics[width=1.0\linewidth]{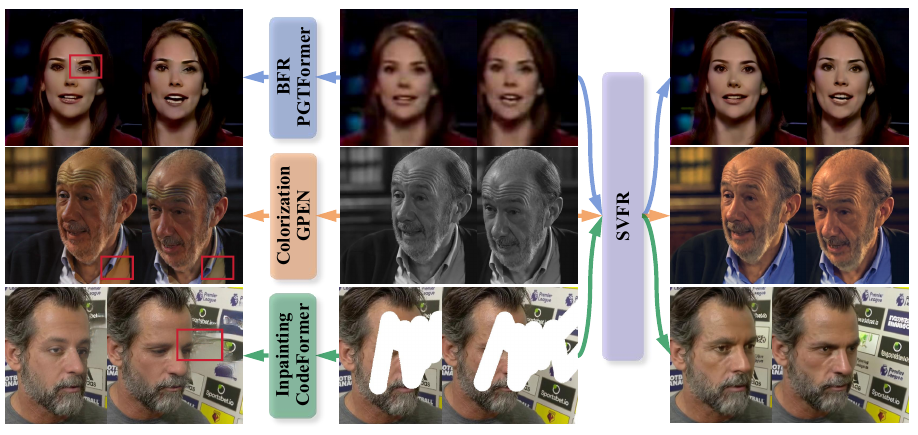}

   \caption{\noindent\textbf{Face restoration results of SOTA single task methods and SVFR.} Our unified SVFR model outperforms SOTA single task methods by avoiding facial structure abnormalities, temporal color inconsistencies, and texture distortions.
   }
   \label{fig:shoutu}
   \vspace{-0.2in}
\end{figure}
% \fi

\section{Introduction \label{sec:label}}

Face Restoration (FR) is a critical area of research within the broader field of image and video processing, aiming to recover high-quality facial images from degraded inputs. This task has significant applications for both images and videos. For example, effective image FR methods can help on personal photo enhancement, historical photo restoration, and forensic analysis, while video FR can well serve the goal of film restoration and surveillance.

\footnotetext[1]{Equal Contribution. †Corresponding Author.}
%\footnotetext[2]{Corresponding Author.}

While abundant works have been proposed to solve image FR with acceptable results, the video FR, which targets high-quality and realistic face recovery with appropriate identity consistency together with temporal stability, has hardly been explored. Few efforts dedicated to this area such as BasicVSR++~\cite{chan2022basicvsr++} and KEEP~\cite{xue2024human} still significantly lack behind their image-based counterparts, suffering from problems such as low generation quality, temporal jitter and discontinuity, due to the multifaceted challenges. The main problem lies in the fact that video FR is intrinsically more complex than image FR in that the temporal dimension introduces additional complexity, requiring the model to maintain consistency across frames while dealing with motion artifacts, occlusions, and varying lighting conditions. However, the high-quality training data for video FR is hard to collect. In this way, the scarcity of high-quality annotated video datasets further complicates the training process, making it difficult to achieve robust performance. Moreover, the limited capacity of architecture adopted by previous works also contributes to the low fidelity problem.

To this end, in this paper, we explore an alternative solution to the above-mentioned problems, that is, leveraging a unified face restoration framework that simultaneously engages more general restoration supervision other than BFR. Such an idea has been successfully applied to a wide range of deep learning tasks. For example, multi-task learning frameworks have shown significant improvements in natural language processing~\cite{worsham2020multi}, where tasks like translation, summarization, and question answering benefit from shared representations. Meanwhile, one can find that BFR intrinsically shares similar goals with other tasks, \textit{e.g.,} the color of low-quality videos is often destroyed due to compression artifacts, whose restoration also involves performing colorization. Intuitively, a well-learned colorization model should enhance the BFR process by providing more accurate color information. Similarly, inpainting techniques, which aim to fill in missing or corrupted regions in images, can provide valuable priors for restoring occluded or degraded facial features in videos.

To validate this idea, we first conduct several pilot experiments to show that with simple transfer learning, the prior knowledge from BFR, inpainting, and colorization can be beneficial for each other. These experiments demonstrate that leveraging shared representations across these tasks can lead to improved performance in video BFR. For instance, we observed that models pretrained on colorization tasks were better at restoring natural skin tones in low-quality videos, while inpainting models helped in reconstructing occluded facial regions with higher fidelity.

Based on this clue, we in this paper propose a novel framework dubbed as Stable Video Face Restoration (SVFR) as a generalized solver for a composited video FR task named Generalized Video Face Restoration (GVFR), including video BFR~\cite{wang2021towards,li2020blind,zhou2022towards}, inpainting~\cite{zhang2017demeshnet,xu2024personalized,sola2023unmasking}, and colorization~\cite{xu2021denseunet}, to enhance the supervision from limited training data. Our proposed SVFR leverages 
strong motion-level and spatial-level priors embedded in pretrained Stable Video Diffusion (SVD)~\cite{blattmann2023stable} with task-specific information, to ensure the backbone model enjoys enough capacity for producing high-fidelity results. Concretely, the pretrained SVD is assisted with conditions from two sources, \textit{i.e.,} the contaminated task-specific context and the reference identity image, to indicate the target task and provide the spatial details. This basic pipeline is modified with several novelties to benefit GVFR. First, in order to ensure task-specific information can be embedded in the proper latent space, \textit{i.e.,} videos with same content and different forms of degradation should be more similar than videos with different contents, a novel unified face restoration framework featuring the task embedding and the unified latent regularization is employed to constrain the intermediate feature from diffusion UNet. Additionally, to further enhance the human face quality generated by our model, we propose a facial prior learning objective, which guides the model with structure priors represented by face landmarks to smoothly embed the human face structure prior into the pretrained SVD. 
During inference, we leverage a self-referred refinement strategy, in which the generated results are refined by referring to the previously generated frames, in order to improve temporal stability. This whole framework can successfully leverage the complementary strength of BFR, inpainting, and colorization, thus enhancing the temporal coherence of the restored video and achieving superior restoration quality.

In order to validate the effectiveness of our method, we conduct experiments on all three subtasks on VFHQ-test dataset, following the previous works. Extensive results show that by addressing the challenges of these video-based low-level tasks through a unified framework, our method paves the way for more robust and efficient video restoration techniques, with potential applications in fields such as video conferencing, film restoration, and surveillance.

In summary, our contributions can be listed as follows:
\begin{itemize}
    \item We propose a novel framework called Stable Video Face Restoration (SVFR) that unifies video BFR, inpainting, and colorization tasks, leveraging shared representations to enhance supervision from limited training data and improve overall restoration quality.
    \item We introduce key innovations including a novel unified latent regularization to ensure proper embedding of task-specific information, and a facial prior learning objective that incorporates human face priors using face landmarks, thereby enhancing the fidelity of the restored videos.
    \item We develop a self-referred refinement strategy during inference, which refines generated frames by referring to previously generated ones, significantly improving temporal stability and coherence in the restored videos.

\end{itemize}

\section{Related Works}

\paragraph{Diffusion models in video processing.} The advent of diffusion models in image generation~\cite{ho2020denoising} has catalyzed their application in video processing. Early research primarily extended these models~\cite{ho2022video,zhang2023i2vgen} to video generation by incorporating temporal dynamics. For instance, Video LDM~\cite{blattmann2023videoldm} begins with image pretraining and subsequently integrates temporal layers fine-tuned on video data. VideoCrafter~\cite{chen2023videocrafter} enhances video generation by combining textual and visual features from CLIP~\cite{radford2021clip} within cross-attention layers. VideoComposer~\cite{wang2024videocomposer} uses images as conditional prompts during training to guide video generation. Building on these foundations, researchers have explored additional video processing tasks. VIDIM~\cite{jain2024vidim}, for example, employs cascaded diffusion models for video interpolation, while DiffIR2VR~\cite{yeh2024diffir2vr} adapts pretrained image-based diffusion models for zero-shot video restoration through latent warping. DiffAct~\cite{liu2023diffact} generates action sequences for temporal action segmentation via iterative denoising. 

\paragraph{Blind face restoration.} The field of blind face restoration aims to recover high-quality face images from severely degraded inputs, often captured in uncontrolled settings. Given the structured nature of faces, researchers have successfully incorporated prior information into restoration models. As an important method to address this task, generative priors from pre-trained GANs, such as StyleGAN~\cite{karras2019style, karras2020analyzing}, are leveraged through iterative latent optimization of GAN inversion \cite{gu2020image,menon2020pulse,pan2021exploiting}. Other works such as GLEAN~\cite{chan2021glean}, GPEN~\cite{yang2021gan}, and GFPGAN~\cite{wang2021towards} were proposed to introduce generative prior to enhance restoration quality. Moreover, recent works~\cite{zhou2022codeformer,gu2022vqfr,wang2022restoreformer, hu2020face, zhu2022blind,li2022faceformer} also tried to enhancing robustness to severe degradation by compressing the high quality images into a finite codebook, or introduce structure prior such as $3$D Morphable Model ($3$DMM)~\cite{blanz1999morphable}. Most face restoration methods are image-based ones~\cite{lin2023diffbir,wang2023dr2,yue2024difface}, without focusing on video-level tasks. Few video face restoration methods~\cite{wang2019edvr,xu2024beyond} such as BasicVSR++~\cite{chan2022basicvsrpp} and KEEP~\cite{feng2024keep} propose to extend the image-based models to video tasks. However, these methods still suffer from severe temporal instability and low quality. In comparison, our proposed method can perform much better due to the design of introducing multiple learning tasks, which has hardly been explored before for BFR.

\section{Pilot Study: Can training on multiple video FR tasks help? \label{sec:pilot}}

\begin{table}[htb]
\small
\caption{Pilot study results. For all subtasks, training models based on specific prior knowledge leads to better FID.}
\label{tab:pilot-study}
\centering
\resizebox{0.6\linewidth}{!}{
\begin{tabular}{ccc}
\toprule
GVFR subtask  & FID w/o prior$\downarrow$ & FID w prior$\downarrow$\\
\midrule
BFR  &30.023 & 28.014 \\
\midrule
Color &12.677 & 11.941\\
\midrule
Inpaint &9.907 & 9.089\\
% \midrule

\bottomrule
\end{tabular}
\vspace{-0.2in}
}
\end{table}

In order to validate the efficacy of leveraging supervision from multiple subtasks to enhance the performance of video FR, we setup a straightforward pilot experiment in this section. Concretely, we train GPEN~\cite{yang2021gan} for three tasks including BFR, colorization and inpainting both with and without using pretrained model with specific prior knowledge. The prior knowledge is set as the pretraining on inpainting for BFR and BFR for the other two tasks. The results are presented in Tab.~\ref{tab:pilot-study}, from which we can find that the transfer learning setting consistently outperforms the training-from-scratch setting in all tasks, thus indicating that these three subtasks in the GVFR indeed share similar prior knowledge that can benefit each other.

\section{Methodology}
\label{sec:methods}
\begin{figure*}[t]
  \centering
   \includegraphics[width=0.95\linewidth]{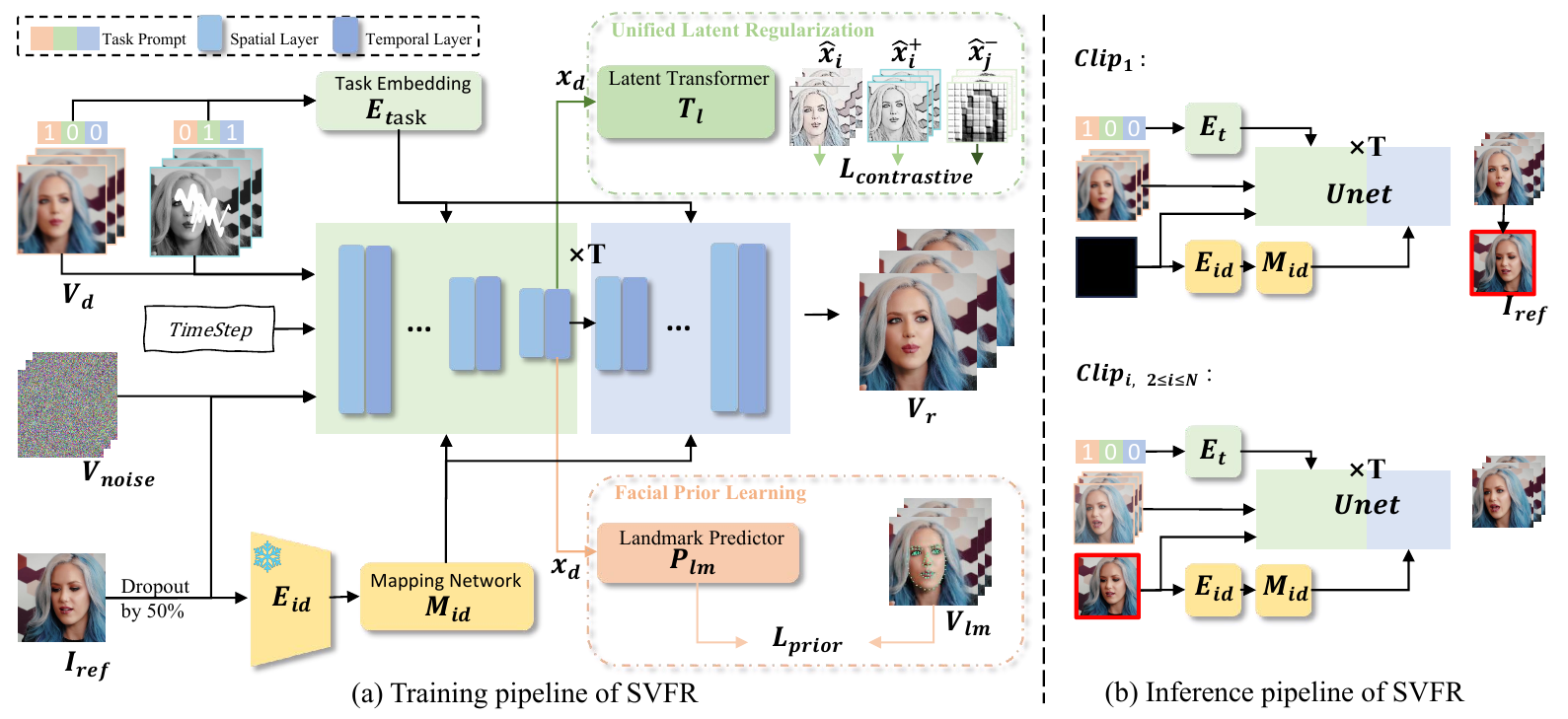}

   \caption{\textbf{Pipeline of our proposed SVFR}. (a) Training pipeline of SVFR: We incorporate task information into the training process using a task embedding approach to enhance unified face restoration. To enhance performance across different tasks, we propose unified latent regularization to effectively aggregate task-specific features, thereby improving the model's performance on various tasks. Additionally, we introduce facial landmark information as an auxiliary guidance to help the model learn structural priors related to human face. Finally, we propose a Self-referred Refinement method that ensures temporal stability. During training, a reference frame is randomly provided, the model will produce consistent results when a reference frame is available. (b) Inference pipeline of SVFR: During inference, we initially generate the first video clip without a reference frame, then select a result frame as the reference image $I_{ref}$ for subsequent video clips.
   }
   \label{fig:ppl}
   \vspace{-0.2in}
\end{figure*}

\noindent\textbf{Preliminary: Stable Video Diffusion.}
Diffusion models~\cite{xu2024beyond} are generative models that use a forward process to turn data \( x_0 \sim p_{\text{data}}(x) \) into Gaussian noise \( x_T \sim \mathcal{N}(0, I) \) and learn a reverse process to remove this noise step-by-step. In the forward process \( q(x_t | x_0, t) \), noise is added gradually over \( T \) time steps, producing increasingly noisy samples \( x_t \). The reverse process, \( p_\theta(x_{t-1} | x_t, t) \), uses a neural network \( \varepsilon_\theta(x_t, t) \) to predict and remove the noise, progressively recovering the original data.

For video generation tasks, several methods are based on diffusion models. In this work, we choose Stable Video Diffusion (SVD)~\cite{blattmann2023svd} as a base model to utilize its time-stable modeling capabilities. To model the temporal data, SVD introduces two types of temporal layers: 3D convolution layers and temporal attention layers. Temporal layers are also integrated into the VAE decoder. Given a video $x \in \mathbb{R}^{N \times 3 \times H \times W} $, we first encode each frame into the latent space, represented as $z=\mathcal{E}(x)$. In this latent space, we perform forward and reverse processes. The final generated video, \( x \), is then obtained by decoding from this latent representation.

\begin{equation}
\label{eq:ldm}
\mathcal{L}_{noise}=\mathbb{E}_{\mathcal{E}(x), \epsilon \sim \mathcal{N}(0,1), t}\left[\left\|\epsilon-\epsilon_{\theta}\left(z^t, t\right)\right\|_2^2\right],
\end{equation}
where $t$ is sampled from $\{0,...,T\}$, $\epsilon_{\theta}$ denotes a diffusion U-Net parameterized by $\theta$. During inference, $z^T$ is directly sampled from Gaussian distribution, which is denoised into $z^0$ and then mapped back to video space using decoder $\mathcal{D}$ of VQ-VAE.

\paragraph{Overview.} To leverage multiple tasks to mutually enhance the performance, we present a unified framework Stable Video Face Restoration (SVFR) as depicted in Fig.~\ref{fig:ppl}. Formally, for each task, the model is provided with a contaminated source video $\mathbf{V}_{d}$ (i.e., low-quality video for BFR, masked video for inpainting, and gray-scale video for colorization), and required to generate a repaired version of it, denoted as $\mathbf{V}_{r}$. In our proposed framework, the task-level conditions are initially processed by a novel unified face restoration framework (Sec.~\ref{sec:task}), where the Task Embedding enhances the model's ability to recognize the specific task, and the Unified Latent Regularization integrates intermediate feature from diverse task domains into a shared latent space. To smoothly embed the human facial prior into the pretrained SVD, we further propose a facial prior learning objective (Sec.~\ref{sec:loss}), which guides the model with structure priors represented by face landmarks. After that, we leverage a self-referred refinement strategy (Sec.~\ref{sec:refine}), in which the generated results are refined by referring to the previously generated frames to improve the temporal stability.

\subsection{Unified Face Restoration Framework \label{sec:task}}

In order to guide the denoising process with source video $\mathbf{V}_{d}$, a simple approach is to encode $\mathbf{V}_{d}$ into latent space and directly concatenate $\mathcal{E}(\mathbf{V}_{d})$ with the noise $z^t$. However, the source videos from different tasks can convey different prior information. For instance, $\mathbf{V}_{d}$ provided by BFR shares the same structural information with $\mathbf{V}_{r}$ while lacking detailed texture information. On the other hand, information from $\mathbf{V}_{r}$ is fully retained in $\mathbf{V}_{d}$ for unmasked regions while being unavailable inside the masked areas. Therefore, simply adopting pretrained VAE cannot encode all source videos into a proper and consistent latent space which can then be used to guide the diffusion U-Net.

To this end, we propose a unified face restoration framework comprising two key modules: Task Embedding and the Unified Latent Regularization. The Task Embedding module is designed to capture the unique characteristics of each task, allowing the model to adjust when switching between tasks adaptively. Meanwhile, the Unified Latent Regularization integrates information across tasks within a shared latent space, ensuring that task-specific features are effectively leveraged for enhanced performance across diverse tasks.

\paragraph{Task embedding.} 
While the source video $\mathbf{V}_{d}$ can indicate the required task to some extent, directly relying on $\mathbf{V}_{d}$ for task information can lead to confusion. For example, both colorization and BFR may involve video with destroyed color information. Moreover, when handling multiple types of contamination, $\mathbf{V}_{d}$ itself cannot provide sufficient information for the target task, thus potentially leading to undesired outcomes.
since $\mathbf{V}_{d}$ from different tasks are input into the model, it is challenging for the model to distinguish the target task without additional guidance, potentially leading to undesired outcomes. 
To this end, we leverage the Task Embedding module as an essential component within the multi-task framework, designed to capture and represent the unique characteristics of each task, thereby enhancing the model's ability to recognize and adapt to specific tasks. 

Given a task set $\mathcal{T}=\{T_{1}, T_{2}, T_{3}\}$, where $T_1$, $T_2$ and $T_3$ represent BFR, colorization, and inpainting tasks respectively. To enhance the model's ability to recognize tasks, each task is represented by a binary indicator $t_i(i \in [1,3])$ (0 for absence and 1 for presence). The task prompt is represented as a binary vector $\gamma = [t_1,t_2,t_3]$. For example, \([0, 1, 1]\) indicates that colorization and inpainting tasks are active while BFR is inactive. For a given task prompt $\gamma$, a Task Embedding layer $\mathbf{E}_{task}$ is employed to map $\gamma$ to a task embedding $\mathbf{E}_{task}(\gamma)$, which is then integrated into the diffusion U-Net to guide task-specific adaptations throughout the model by adding it to the time embedding. 

\paragraph{Unified Latent Regularization.} 
To further align features across different tasks and enable the model to leverage shared knowledge within a unified learning framework, we propose Unified Latent Regularization (ULR). 
ULR constrains the feature distributions of various degraded tasks for the same video within the U-Net feature space, allowing the model to learn common features across multi-tasks.

Specifically, we implement cross-task alignment using features from the middle block of the UNet. We first extract the output from the intermediate layers of the UNet, as these layers capture essential structural and semantic information crucial for modeling nuanced video details. These intermediate outputs $x_d$ are then processed through a Latent Transformer $T_l$, composed with two resblock and a layernorm,  producing feature representations of the video frames $\hat{x}_d$. 

Subsequently, we compute the contrastive loss on these features to improve visual consistency across tasks and reinforce restoration quality. 
This ensures that output frames across tasks share unified intermediate features during restoration, thereby enhancing structural consistency, reducing artifacts, and elevating overall generation quality. Formally, the proposed regularization term for each batch can be represented as:
\begin{equation}
    \mathcal{L}_{ULR} = - \log 
    \left( 
        \frac{\exp \left( (\hat{x}_i \cdot \hat{x}_{i}^{+})/\tau \right)}
        {\sum^{N}_{j=1} \exp \left( (\hat{x}_i \cdot \hat{x}_{j}^{-})/\tau \right)} 
    \right),
\end{equation}
\iffalse
\begin{equation}
    \mathcal{L}_{ULR} = -\frac{1}{N} \sum_{i=1}^{N} \log 
    \left( 
        \frac{\exp \left( \frac{x_i \cdot x_{i}^{+}}{\tau} \right)}
        {\sum^{N}_{j=1} \exp \left( \frac{x_i \cdot x_{j}^{-}}{\tau} \right)} 
    \right),
\end{equation}
\fi
where $\hat{x}_i$ denotes the intermediate features obtained from the Latent Transformer, representing the primary degraded video. $\hat{x}_{i}^{+}$ indicates features derived from different degradations of the same source video, while $\hat{x}_{j}^{-}$ corresponds to features from different source videos, also extracted through the UNet and transformer module. By applying this regularization constraint, ULR effectively maintains spatial and detail feature consistency across tasks, thereby enhancing fidelity.

\subsection{Facial Prior Learning \label{sec:loss}}
While the pretrained SVD can be directly adapted to our task by optimizing the noise prediction loss as in Eq.~\ref{eq:ldm}, such an objective cannot help inject the structure priors of human faces into the model. As a result, the model may struggle to generate facial details with consistent structure.

To address this problem and further guide the model in preserving the facial structure, we introduce an auxiliary objective function that can encourage the model to concentrate on the human face structure-related information from noisy latent. Concretely, we leverage the features $x_d$ output from the U-Net middle block and pass them through a landmark predictor $P_{lm}$ composed of an average pooling layer followed by five layers of MLP. This predictor is trained to estimate a 68-points facial landmark set, obtained from ground truth frames using a pretrained landmark detection model~\cite{guo2018stacked}.

To achieve precise alignment between predicted and ground-truth landmarks, inspired by~\cite{feng2018wing}, we utilize a facial prior loss for fine-grained facial landmark regression. Formally, for each predicted landmark $\hat{y}_i=P_{lm}(x_d)$ and its corresponding ground truth $y_i$ from $V_{lm}$, this loss function is defined as:
\begin{equation}
    \mathcal{L}_{prior} = \left\{\begin{matrix}  
    w\mathrm{ln} (1+ \left | x \right | / \epsilon) & \mathrm{if}  \left | x \right | < w \\  
    \left | x \right | - C & \mathrm{otherwise} 
    \end{matrix}\right.,
    \label{eq:wing_loss-1}
\end{equation}
where $x$ denotes the difference $\hat{y}_i - y_i$, with $w$ and $\epsilon$ as hyperparameters controlling the balance between linear and logarithmic scaling, and $C$ ensuring continuity. 
\iffalse
\begin{equation}
    \mathcal{L}_{prior-overall} = \frac{1}{N} \sum_{i=1}^{N} \mathcal{L}_{prior}(\hat{y}_i - y_i),
    \label{eq:wing_loss-2}
\end{equation}
\fi
This loss is incorporated as an auxiliary component in the total loss function, weighted to balance with other objectives, thereby guiding the model to prioritize facial structure consistency across different tasks. This approach selectively emphasizes smaller deviations while maintaining robustness to outliers, making it highly effective for accurate facial feature alignment. 

In all, the diffusion U-Net $\epsilon_{\theta}$ is optimized together with the unified latent regularization $T_l$, the task embedding $E_{task}$ and the landmark predictor $P_{lm}$, using the following composited objective function:
\begin{equation}
    \mathcal{L}=\mathcal{L}_{noise} + \lambda_1 \mathcal{L}_{ULR} + \lambda_2 \mathcal{L}_{prior},
\end{equation}
where $\lambda_1$, $\lambda_2$ is a hyper-parameter. The detailed model architecture is provided in the supplementary materials.

\subsection{Self-referred Refinement \label{sec:refine}}
To achieve temporal stability and consistency across entire segments in long video facial restoration tasks, we propose a mechanism called Self-referred Refinement. This method enhances generation quality by injecting reference frame features during the generation process, ensuring structural and stylistic consistency between frames and improving overall generation results.

In the training phase, we extract features from the reference frame $\mathbf{I}_{ref}$ using a VAE encoder $z_{ref} = \mathcal{E}(\mathbf{I}_{ref})$. These features are then injected into the initial noise of the U-Net model. Additionally, identity features $\mathbf{f_{id}}$ are extracted by $E_{id}$ and passed through a mapping network $M_{id}$ before being injected into the cross-attention layers of the U-Net. To improve model generalization and allow the model to generate content without a reference frame, we apply a 50\% dropout to the reference frame during training. This strategy encourages the model to develop independent content restoration capabilities, enabling it to produce consistent facial structures even without explicit conditioning.

In inference, for sequential generation across video segments, the model first generates the initial video segment without a reference frame. From this generated segment, we select one frame as a reference for subsequent segments. This approach ensures stylistic and structural continuity across frames, facilitating long-range consistency throughout the entire video sequence.

\begin{table*}[t]
\small
\caption{\noindent \textbf{Quantitative comparisons with state-of-the-art methods on the VFHQ-test dataset.}
 The results show that our method achieves the best performance across both image and video metrics, outperforming all methods on the GVFR tasks. Notably, our approach is tested on a unified model, while other methods use specially trained model for each task. PGTFormer~\cite{xu2024beyond} and KEEP~\cite{feng2024keep} did not provide trained models for colorization and inpainting, so the metrics for these two tasks are empty.}
   \vspace{-0.1in}
\label{tab:compare-sota-method}
\centering
\resizebox{0.98\textwidth}{!}{
\begin{tabular}{cccccccc}
\toprule
\multirow{2}{*}{Methods} & \multicolumn{7}{c}{BFR / Colorization / Inpainting}  \\
\cline{2-8} 
& PSNR↑ & SSIM↑ & LPIPS↓ & IDS↑ & VIDD↓ & FVD↓  \\
\midrule
GPEN~\cite{yang2021gan} &26.237 / \underline{22.012} / 23.596 &0.795 / \underline{0.887} / 0.886 &0.320 / \underline{0.294} / 0.228 &0.786 / \underline{0.959} / 0.757 &\underline{0.575} / \underline{0.499} / 0.671 &412.81 / \underline{366.2481} / 286.763 \\
CodeFormerr~\cite{zhou2022codeformer} &26.528 / 17.927 / \underline{28.672} &\underline{0.762} / 0.663 / 
\underline{0.896} &0.361 / 0.508 / \underline{0.156} &0.784 / 0.624 / \underline{0.846} &0.700 / 0.895 / 0.727 &379.53 / 551.448 / \underline{170.631} \\
PGDiff~\cite{yang2024pgdiff} &23.638 / 21.550 / 20.551 &0.765 / 0.825 / 0.794 &0.399 / 0.481 / 0.360 &0.489 / 0.794 / 0.615 &0.649 / 0.558 / \underline{0.595} &445.36 / 370.150 / 354.615 \\
\midrule
PGTFormer~\cite{xu2024beyond} &\underline{28.996} / - / - &0.843 / - / - &\underline{0.248} / - / - &\underline{0.845} / - / - &0.577 / - / - &\underline{154.857} / - / - \\
KEEP~\cite{feng2024keep} &27.335 / - / - &0.813 / - / - &0.259 / - / - &0.790 / - / - &0.787 / - / - &399.239 / - / - \\
Ours &\textbf{29.563} / \textbf{23.079} / \textbf{29.119}  &\textbf{0.862} / \textbf{0.896} / \textbf{0.904}  &\textbf{0.223} / \textbf{0.272} / \textbf{0.153} & \textbf{0.902} / \textbf{0.980} / \textbf{0.888} & \textbf{0.479} / \textbf{0.497} / \textbf{0.504} & \textbf{89.316} / \textbf{204.260} / \textbf{88.354} & \\
\bottomrule
\end{tabular}
}
\vspace{-0.25in}
\end{table*}

\section{Experiments}
\subsection{Implementation detail \label{sec:experiment-detail}}

\noindent\textbf{Dataset.} We construct the training datasets by following previous works focusing on three subtasks. Specifically, We selected VoxCeleb2~\cite{chung2018voxceleb2}, CelebV-Text~\cite{yu2023celebv}, and VFHQ~\cite{xie2022vfhq} as our training datasets.
%To enhance the quality of our training data, we filtered these datasets rigorously. We first extracted square face bounding boxes and scaled them by a factor of 0.2. We crop and filter out those with resolutions below 512. Subsequently, we applied the image quality assessment method ARNIQA~\cite{agnolucci2024arniqa}, using a model trained on "live" data, to further select frames with scores above 0.75.
% This process yielded a high-quality video dataset comprising 20,000 clips.
To enhance the quality of our training data, we utilize ARNIQA~\cite{agnolucci2024arniqa} score to filtere a high-quality video dataset comprising 20,000 clips.
The VFHQ-test dataset~\cite{xie2022vfhq} is selected as test dataset, which contains 50 videos. For the BFR task, we applied various degradations, including random blurring, noise addition, resizing, and video compression (CRF). For the colorization task, we converted the videos to grayscale, and for the inpainting task, we applied brush stroke masks~\cite{yang2021gan} to the video frames. Since some comparison methods operate only on aligned images or videos, we performed alignment on the test data.

\noindent\textbf{Metrics.} To evaluate our model’s performance in face restoration, we use a range of targeted metrics. PSNR and SSIM~\cite{wang2004image} assess frame-level reconstruction accuracy, while LPIPS ~\cite{zhang2018unreasonable} captures perceptual quality. IDS evaluates identity consistency in face restoration. At the video level, VIDD~\cite{chen2024towards} measures temporal consistency by calculating identity feature differences between adjacent frames using ArcFace, and FVD~\cite{unterthiner2019fvd} assesses overall video quality, reflecting spatial and temporal coherence.

\noindent\textbf{Baseline.} For the BFR task, we benchmarked our approach against state-of-the-art methods, including image-based methods such as GPEN~\cite{yang2021gan}, CodeFormer~\cite{zhou2022codeformer}, and PGDiff~\cite{yang2024pgdiff}, which generate frames independently, processing them one by one. We also compared with video-based restoration methods like PGTFormer~\cite{xu2024beyond} and KEEP~\cite{feng2024keep}. All testing degradations were applied consistently across methods, and each baseline followed the settings reported in their respective original papers. For the inpainting and colorization tasks, we selected GPEN~\cite{yang2021gan}, CodeFormer~\cite{zhou2022codeformer}, and PGDiff~\cite{yang2024pgdiff} for comparison.

\subsection{Comparison with State-of-the-Art Methods}
\noindent\textbf{Qualitative analysis.}
In Table \ref{tab:compare-sota-method}, we present a comprehensive comparison of our method against state-of-the-art techniques for the BFR, colorization, and inpainting tasks on the VFHQ-test dataset.
Our approach outperforms existing methods across a range of image quality metrics, including PSNR, SSIM and LPIPS, showcasing its superior ability to restore high-quality video frames. Notably, the higher IDS score demonstrates our method’s enhanced capacity to preserve facial identity consistency throughout the restoration process. Furthermore, the improved VIDD and FVD scores reflect significant gains in both the realism and temporal stability of the restored video, underscoring the robustness of our method in handling dynamic video restoration tasks.

\begin{figure*}[t]
  \centering
   \includegraphics[width=0.93\linewidth]{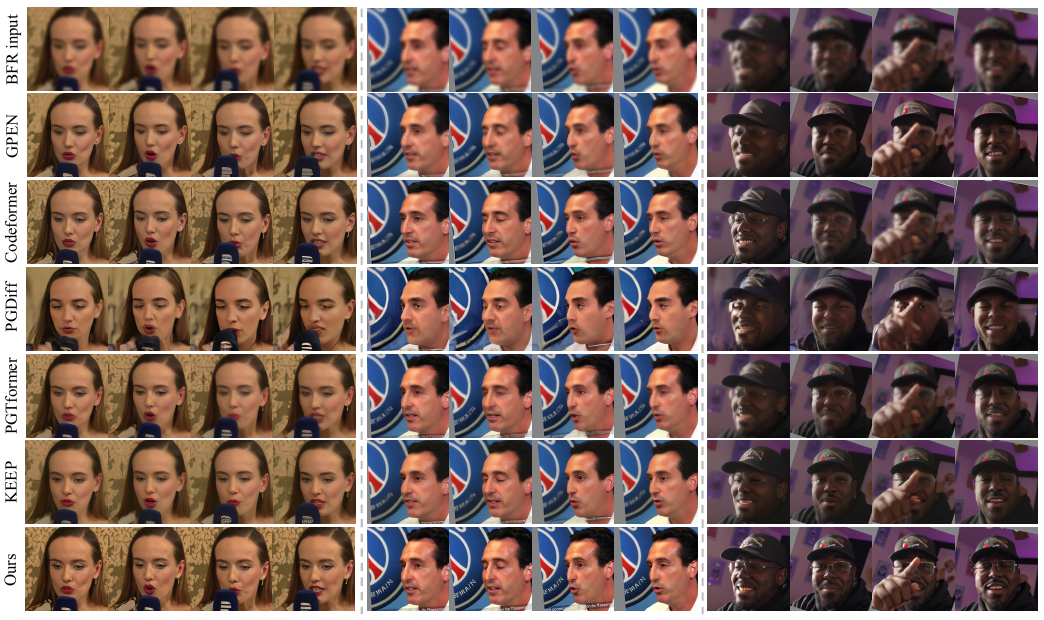}
   \vspace{-0.1in}
    \caption{\noindent \textbf{Comparisons of SVFR with Advanced BFR Methods.} Compared to other methods, SVFR avoids distortions in facial structure and shifts in facial attributes, resulting in highly realistic faces. It also demonstrates high spatial and temporal stability in complex scenarios such as occlusion, side-profile views, and large motion.}

   \label{fig:visual-1}
   \vspace{-0.1in}
\end{figure*}

\begin{figure*}[htb]
  \centering
   \includegraphics[width=0.93\linewidth]{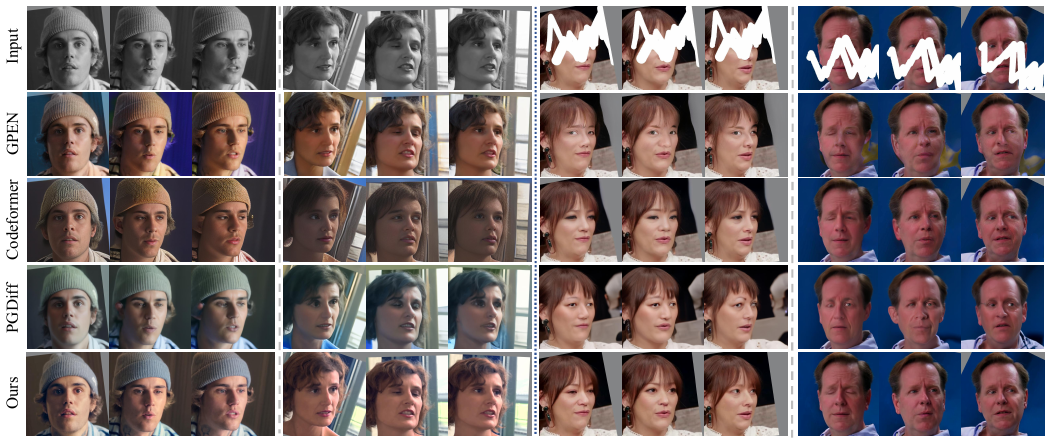}

   \vspace{-0.1in}
    \caption{\noindent \textbf{Comparisons of SVFR with Advanced Colorization (left) and Inpainting (right) Methods.} Compared to other methods, our model preserves realistic facial structure and attributes, while keeping detail preservation, and cross-frame stability across extended sequences.}
   \vspace{-0.1in}

   \label{fig:visual-inpainting}
   \vspace{-0.1in}
\end{figure*}

\noindent\textbf{Quantitative analysis.}
To further validate the effectiveness of our approach, we performed a visual comparison with existing methods, as shown in Fig. \ref{fig:visual-1}. The results demonstrate that other methods often struggle with issues such as frame-level blurring, facial feature distortions, and producing artifacts,  especially when there are large pose or motion changes between consecutive frames. 
Additionally, these methods tend to fail in challenging scenarios, including facial occlusion, the presence of accessories such as glasses or earrings, and side-profile views, where our approach consistently delivers more accurate and stable restorations. 
For instance, in the first case, when the face is obstructed by a microphone, other methods tend to distort the mouth region during generation, thereby compromising the overall facial structure in the restored video.
Meanwhile, they also suffer from temporal inconsistencies, including abrupt changes in facial features, flickering, and poor frame-to-frame continuity, which degrade the overall stability and realism of the restored video.

Additionally, we conducted a visual comparison for the colorization and inpainting tasks. Fig.\ref{fig:visual-inpainting}(left) presents the results for the colorization task, where our method produces clear, faithful video frames that are consistent with the input grayscale video. Unlike CodeFormer and PGDiff, our approach avoids issues such as distorted facial attributes (e.g., unrealistic eyes, etc.). Furthermore, leveraging the Self-referred Refinement strategy, our method does not introduce color shifts across frames, maintaining temporal consistency. Fig.\ref{fig:visual-inpainting}(right) presents the comparison results for the inpainting task. The results show that other methods tend to generate distorted or collapsed facial structures in the masked regions, especially when head pose transformations occur. In contrast, our method effectively reconstructs facial features with high accuracy, preserving both structural integrity and consistency across frames.

\begin{figure}[htb]
  \centering
   \includegraphics[width=1.0\linewidth]{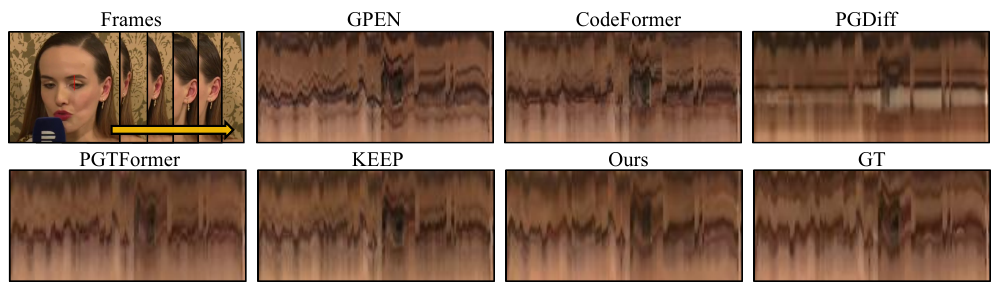}

   \vspace{-0.1in}
   \caption{\textbf{Temporal Stability Comparison.} Through temporal stacking visualizations, we observe that our method exhibits significantly less jitter compared to other approaches, demonstrated the superior temporal stability of our approach, ensuring that the restored video frames maintain both spatial coherence and temporal continuity, even across long sequences.
   }
   \vspace{-0.1in}
   \label{fig:jitter}
\end{figure}

To assess temporal stability, we utilize a temporal stacking visualization, where a vertical line of pixels from the same spatial location is extracted across consecutive frames. As shown in Fig.\ref{fig:jitter}, image-based methods often suffer from jitter and misalignment along the temporal axis, while video-based methods exhibit limited texture continuity. In contrast, our approach ensures a smooth and consistent transition across frames, as evidenced by the seamless and coherent texture in the stacked visualization. This underscores the robustness of our method in preserving both facial feature integrity and temporal stability throughout the video sequence.

\iffalse

\begin{table}[htb]
\small
\caption{Quantitative comparisons with state-of-the-art colorization methods on VFHQ-test dataset.}
\label{tab:colorization}
\centering
\resizebox{\linewidth}{!}{
\begin{tabular}{cccccccc}
\toprule
Methods & PSNR↑ & SSIM↑ & LPIPS↓ & FID↓ & IDS↑ & VIDD↓ & FVD↓ \\
\midrule
GPEN~\cite{yang2021gan} & & & & & & & \\
CodeFormerr~\cite{zhou2022codeformer} & & & & & & & \\
PGdiff~\cite{yang2024pgdiff} & & & & & & & \\
\midrule
Ours & & & & & & & \\
\bottomrule
\end{tabular}
}
\end{table}

\noindent\textbf{Inpainting.}

\begin{table}[htb]
\small
\caption{Quantitative comparisons with state-of-the-art inpainting methods on VFHQ-test dataset.}
\label{tab:inpainting}
\centering
\resizebox{\linewidth}{!}{
\begin{tabular}{cccccccc}
\toprule
Methods & PSNR & SSIM & LPIPS & FID & IDS & VIDD & FVD \\
\midrule
GPEN~\cite{yang2021gan} & & & & & & & \\
CodeFormerr~\cite{zhou2022codeformer} & & & & & & & \\
PGdiff~\cite{yang2024pgdiff} & & & & & & & \\
\midrule
Ours & & & & & & & \\
\bottomrule
\end{tabular}
}
\end{table}

\fi

\begin{table*}[htb]
\small
\caption{Ablation study for multi-task training.}
   \vspace{-0.1in}
\label{tab:ablation}
\centering
\resizebox{0.98\textwidth}{!}{
\begin{tabular}{cccccccc}
\toprule
\multirow{2}{*}{Methods} & \multicolumn{7}{c}{BFR / Colorization / Inpainting}  \\
\cline{2-8} 
& PSNR↑ & SSIM↑ & LPIPS↓ & IDS↑ & VIDD↓ & FVD↓  \\
\midrule
single &28.323 / 22.233 / 27.765 &0.845 / 0.831 / 0.896 &0.260 / 0.279 / 0.162 &0.854 / 0.978 / 0.861 &0.495 / 0.503 / 0.517 &167.313 / 233.389 / 106.522 \\
multi &28.936 / 22.921 / 28.303 &0.854 / 0.870 / 0.898 &0.242 / 0.274 / 0.156 &0.881 / \underline{0.979} / 0.875 &0.489 / 0.501 / 0.511 &98.781 / 223.168 / 101.146 \\
multi+ULR &\underline{29.296} / \underline{22.987} / \underline{28.337}  &\underline{0.859} / \underline{0.886} / \underline{0.900}  &\underline{0.225} / \textbf{0.270} / \underline{0.155} &\underline{0.884} / 0.978 / \underline{0.879} &\underline{0.486} / \underline{0.498} / \underline{0.508} &\underline{90.353} / \underline{214.846} / \underline{93.615} \\
\midrule
multi+ULR+FPL &\textbf{29.563} / \textbf{23.079} / \textbf{29.119}  &\textbf{0.862} / \textbf{0.896} / \textbf{0.904}  &\textbf{0.223} / \underline{0.272} / \textbf{0.153} &\textbf{0.902} / \textbf{0.980} / \textbf{0.888} &\textbf{0.479} / \textbf{0.497} / \textbf{0.504} &\textbf{89.316} / \textbf{204.260} / \textbf{88.354}  & \\
\bottomrule
\end{tabular}
}
   \vspace{-0.2in}
\end{table*}

\subsection{Ablation Study}

To further verify the efficacy of our contributions, we conduct several ablation studies with evaluation on VFHQ test dataset. Due to space limitations, additional ablation studies are provided in the supplementary materials.

\noindent\textbf{Multi-Task training.} As demonstrated in Sec.~\ref{sec:pilot}, three subtasks in our proposed GVFR share common prior knowledge. To further validate that training on GVFR improves performance compared to training each task individually, Tab.~\ref{tab:ablation} compares the performance metrics of each task when trained separately versus within the GVFR framework. The results clearly show that multi-task training enhances overall model performance by facilitating the transfer of shared features across tasks. This leads to better detail preservation and temporal stability, confirming that joint training is beneficial for comprehensive facial video restoration.

\iffalse
\begin{table}[htb]
\small
\caption{Ablation study for Unified Face Restoration Framework.}
\label{tab:ablation-module}
\centering
\resizebox{\linewidth}{!}{
\begin{tabular}{cccccccc}
\toprule
variants & PSNR & SSIM & LPIPS & FID & IDS & VIDD & FVD \\
\midrule
naive & & & & & & & \\
+task emb. & & & & & & & \\
+task emb. + ULR & & & & & & & \\
\midrule
+task emb. + ULR + FPL & & & & & & & \\
\bottomrule
\end{tabular}
}
\end{table}
\fi

\noindent \textbf{Effectiveness of Unified Latent Regularization.}
We conduct ablation studies on Unified Latent Regularization (ULR) with $\lambda_1 = 0.01$, In Tab.~\ref{tab:ablation}, the results show that ULR improves consistency across multiple restoration tasks by facilitating the sharing of common features and ensuring coherence within frames from the same source. 
By optimizing the shared feature representation in multi-task learning, ULR not only preserves facial characteristics but also enhances cross-task knowledge transfer, enabling better adaptation to diverse restoration challenges.This unified framework significantly improves video clarity, refines facial details, and strengthens temporal stability, ensuring smooth frame transitions and reduced artifacts. Ultimately, ULR enables tasks to mutually reinforce each other, leading to robust and comprehensive facial video restoration.

\noindent\textbf{Effectiveness of facial prior learning.} As shown in Tab.~\ref{tab:ablation} with $\lambda_2 = 0.1$, $\mathcal{L}_{prior}$ enhances temporal consistency and ensures spatial coherence, particularly in BFR and inpainting, even under challenging conditions such as occlusions and side-profile views. By aligning facial features structurally, the prior loss significantly improves model stability and robustness in video restoration.

\begin{figure}[t]
  \centering
   \includegraphics[width=1.0\linewidth]{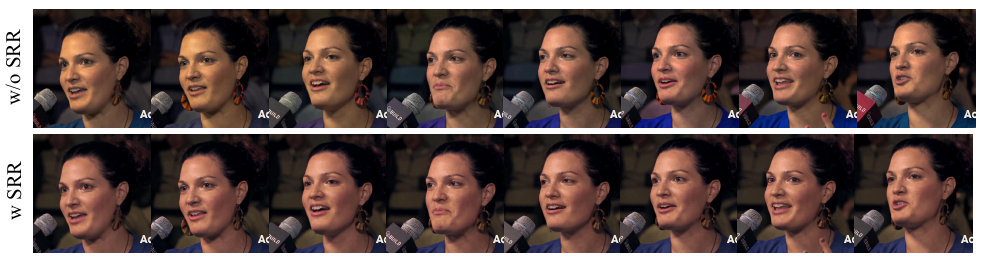}

   \vspace{-0.1in}
   \caption{\noindent\textbf{Ablation study on self-referred refinement (SRR).} Comparison of model inference with and without the SRR strategy shows that incorporating SRR results in more stable temporal consistency, including consistent facial attributes and video color.
   }
   \label{fig:ablation-srr}
   \vspace{-0.1in}
\end{figure}

\noindent\textbf{Effectiveness of self-referred refinement.}  To validate the efficacy of the proposed self-referred refinement(SRR), we provide visual comparisons of inference results with and without the Self-referred Refinement strategy. As observed in Fig.\ref{fig:ablation-srr}, without SRR, inconsistencies emerge over long video sequences, such as color shifts in colorization tasks and facial misalignment in inpainting tasks, especially in redrawn regions. By contrast, incorporating SRR allows the model to maintain consistent identity, color, and facial attributes across extended sequences. This refinement mechanism reinforces temporal stability, ensuring that features remain cohesive throughout the video, thereby greatly enhancing the reliability and continuity of restored facial attributes in long-sequence video restoration.

\iffalse
\begin{table}[htb]
\small
\caption{Ablation study for facial prior learning. For all metrics, the larger score denotes the better model.}
\label{tab:ablation-loss}
\centering
\resizebox{\linewidth}{!}{
\begin{tabular}{cccccccc}
\toprule
variants & PSNR & SSIM & LPIPS & FID & IDS & VIDD & FVD \\
\midrule
w/o $\mathcal{L}_{prior}$ & & & & & & & \\
w $\mathcal{L}_{prior}$ & & & & & & & \\
\bottomrule
\end{tabular}
}
\end{table}
\fi

\iffalse
\begin{table}[htp]
    \centering
    \caption{caption caption caption
    }
    \label{tab:xxx}
    \renewcommand{\arraystretch}{1.1}
    \setlength\tabcolsep{5.0pt}
    \resizebox{0.9\linewidth}{!}{
        \begin{tabular}{ccccc}
        \toprule[0.1em]
        caption & caption & caption & caption & caption \\
        \hline
        1 & 2 & 3 & 4 & 5 \\
        1 & 2 & 3 & 4 & 5 \\
        \hline
        1 & 2 & 3 & 4 & 5 \\
        1 & 2 & 3 & 4 & 5 \\
        \toprule[0.1em]
        \end{tabular}
    }
\end{table}
\fi

\section{Conclusion}

In this paper, we have introduced the Generalized Video Face Restoration (GVFR), which is an extended task composed of video BFR, inpainting and colorization. A novel unified face restoration framework is designed to simultaneously address all three challenges in videos. By engaging supervision from all three tasks rather than training models for each task respectively, our method enhances the overall quality and temporal consistency of restored videos. Our approach integrates representative motion priors from pretrained SVD with task-specific information through a unified latent regularization and facial prior learning, resulting in significant improvements over existing state-of-the-art methods. Extensive experimental evaluations demonstrate the efficacy of our method in achieving superior restoration quality and temporal coherence.  

{
    \small
    \bibliographystyle{ieeenat_fullname}
    \bibliography{main}
}


\end{document}